\documentclass{article}
\usepackage[utf8]{inputenc}
\usepackage{textgreek}
\usepackage{graphicx}
\usepackage{subcaption}
\usepackage{booktabs}
\usepackage{amsmath}

\usepackage[dblblindworkshop, final]{neurips_2026}

\workshoptitle{Foundation Models for the Brain and Body}

\usepackage[utf8]{inputenc} 
\usepackage[T1]{fontenc}    
\usepackage{hyperref}       
\usepackage{url}            
\usepackage{booktabs}       
\usepackage{amsfonts}       
\usepackage{nicefrac}       
\usepackage{microtype}      
\usepackage{xcolor}         

\title{Benchmarking EEG Foundation Models for Decoding Alzheimer’s Disease–Relevant Neural Dynamics}

\author{
Konstantinos Barmpas\textsuperscript{*} \\
Imperial College London \& Cogitat \\
\texttt{konstantinos.barmpas16@imperial.ac.uk} \\
\And
Na Lee \textsuperscript{*} \\
Imperial College London \& Cogitat \\
\texttt{na.lee12@ic.ac.uk} \\
\AND
Yannis Panagakis \\
National and Kapodistrian University of Athens \& \\Archimedes Research Unit \& Cogitat \\
\And
Nikolaos Laskaris \\
Aristotle University of Thessaloniki \& Cogitat \\
\And
Shibeshih Mitiku Belachew \\
Indivi AG \\
\AND
Dimitrios A. Adamos \\
Imperial College London \& Cogitat \\
\And
Stefanos Zafeiriou \\
Imperial College London \& Cogitat \\
}

\begin{document}

\maketitle

\renewcommand{\thefootnote}{\fnsymbol{footnote}} 
\footnotetext[1]{These authors contributed equally to this work.}

\begin{abstract}
Early detection and measurement of Alzheimer’s disease (AD) has the potential to delay disease progression, yet existing fluid-based and imaging diagnostic and monitoring approaches are often costly and typically used only after clinical symptoms appear. Electroencephalography (EEG) offers a low-cost, non-invasive and portable alternative, but substantial inter-subject variability has limited its potential as a practical screening tool. EEG foundation models (EEG-FMs), pre-trained on large-scale unlabeled datasets, offer the prospect of learning representations that generalize across individuals. However, their effectiveness on AD-related tasks remains largely unexplored. In this work, we benchmark six state-of-the-art EEG-FMs against a deep learning model trained from scratch. Using a minimal preprocessing pipeline and subject-independent cross-validation, we evaluate model performance across four levels of difficulty: basic brain states (eyes open/closed), AD detection (classification between AD and control), AD-relevant cognitive task dynamics (executive function, attention and short-term/working memory) and subtle pre-symptomatic genetic risk associated with the ApoE-$\varepsilon 4$ allele. We consider both full fine-tuning and head-only settings. Several EEG-FMs recover brain states and working-memory load, demonstrating strong transferability. Several models also retain these representations under head-only training. In contrast, prediction of genetic risk remains at chance level across all evaluated models. These findings establish that EEG-FMs can capture AD-relevant functional and pathological signals across unseen individuals and support their further evaluation as scalable biomarkers of AD diagnosis and disease progression in early-stage AD trials.
\end{abstract}

\section{Introduction}

Alzheimer's disease (AD) is a progressive neurological disorder that primarily affects older adults. AD is characterized by a decline of a person's cognitive functions, including memory, perception and decision-making ~\citep{molecules25245789}. 
Current AD detection methods such as amyloid-beta quantification on Positron Emission Tomography (PET) ~\citep{Liu2026} or emerging efforts using Magnetic Resonance Imaging (MRI) ~\citep{Ma2026} require specialized clinical expertise and due to their high cost, patients may not undergo these diagnostic procedures until the disease has progressed and overt symptoms have been developed. With disease-modifying therapies now available ~\citep{Cummings2023}, early AD identification especially at the pre-symptomatic stage has direct clinical value.

Electroencephalography (EEG) is a non-invasive and cost-effective technique to measure unique neural signatures of  brain activity. EEG can provide real-time measurements of the subject-specific ground truth of impairment in neural dynamics which is known to be present at the earliest and covert stages of the pathological continuum of AD. Complex EEG-derived signatures may also help disentangling the biological heterogeneity of the disease. Over the years, EEG-based AD detection methods have primarily relied on either handcrafted biomarkers ~\citep{Wang2017, Azami2019}, such as phase shift and power spectral density or data-driven deep learning approaches ~\citep{RonceroParra2024}. However, as in many EEG-related applications, individual differences in brain activity (known as inter-subject variability) ~\citep{Barmpas_2024_causality_jne, myICRL} often prevent these features from generalizing effectively to unseen patients and real-world settings, limiting the practicality of EEG-based methods as an everyday AD detection and measurement tool.

In recent years, EEG foundation models (EEG-FMs) have emerged inspired by the success in other computer science fields like language
~\citep{brown2020language, touvron2023llama} and computer vision~\citep{mizrahi2023m}. These EEG foundation models pre-train on large unlabelled corpora via self-supervision, aiming to generalise across subjects, tasks and conditions with minimal labelled fine-tuning data. Although several EEG-FMs have been proposed ~\citep{NEURIPS2023_f6b30f3e, LaBraM, wang2025cbramod, ouahidi2025reve, LEAD, NeuroRVQ} and have demonstrated competitive performance on various downstream tasks~\citep{lee2025largebrainwavefoundationmodels}, their potential exploration for AD-related tasks remain scarce ~\citep{lin2026leveragingfoundationmodeleegbased}.

In this work, using two AD-related datasets, namely PEARL-Neuro ~\citep{PEARL} and Miltiadous2023 ~\citep{Miltiadous2023}, we examine whether the promising scalable state-of-the-art EEG-FMs can extract \emph{(i)} brain states (eyes open / closed), \emph{(ii)} cognitively meaningful task dynamics,  \emph{(iii)} classification between AD and controls, and \emph{(iv)} the genetic risk of AD associated with ApoE-ε4 carrier status in cognitively healthy, middle-aged adults.

\section{Background}
\label{sec:background}

\subsection{General EEG Foundation Models}
In recent years, several EEG foundation models (EEG-FMs) have been developed, differing substantially in their positional encoding schemes, pre-training objectives and tokenisation strategies. 

BIOT ~\citep{NEURIPS2023_f6b30f3e} was among the first foundation models developed for biosignals and employs contrastive pre-training. LaBraM ~\citep{LaBraM} segments EEG recordings into channel patches and uses a vector-quantised (VQ) neural tokeniser to construct a codebook of neural representations. A transformer-based model is then pre-trained to predict masked tokens from this VQ codebook. CBraMod ~\citep{wang2025cbramod} 
uses a criss-cross transformer for separate spatial and temporal channel attention and asymmetric convolutional positional encoding. NeuroRVQ ~\citep{NeuroRVQ} extends LaBraM’s VQ-based approach to multiple frequency scales and separate residual vector-quantised (RVQ) codebook per scale for finer-details. Combined with a phase- and amplitude-aware training objective, the NeuroRVQ tokeniser achieves high-fidelity signal reconstruction across different frequency bands. REVE ~\citep{ouahidi2025reve} uses a four-dimensional positional encoding setting that computes embeddings from the three-dimensional electrode coordinates and temporal patch indices. It is currently the EEG-FM with the largest pre-training corpus.

\subsection{AD-Specific EEG Foundation Models}
In contrast to the above mentioned EEG-FMs, which are pre-trained on diverse and heterogeneous EEG corpora, LEAD ~\citep{LEAD} is the first EEG foundation model specifically designed for Alzheimer’s disease (AD) detection. It is pre-trained using a medical contrastive learning framework and fine-tuned on independent AD datasets. LEAD incorporates a gated temporal-spatial transformer capable of handling variations in recording duration, channel configuration and sampling rate. 

\section{Method}

\subsection{Tasks}

All EEG-FM models described in Section \ref{sec:background} were evaluated in tasks described in PEARL-Neuro ~\citep{PEARL} and Miltiadous2023 ~\citep{Miltiadous2023} datasets.

In PEARL-Neuro ~\citep{PEARL}, each participant (N=79, aged 50–63) completed a resting-state protocol and two cognitive tasks paradigms: the Multi-Source Interference Task (MSIT) designed to measure executive function and attention and the Sternberg memory task, which stimulates and evaluates short-term/working memory. Because the PEARL-Neuro cohort
is enriched in subjects carrying a genetic risk of AD, it supports the study of pre-symptomatic AD markers through the condition-level effects within each cognitive task and the potential decoding of a specific (ApoE-ε4 allele carrier status) genetic risk in the absence of overt disease. In the PEARL-Neuro dataset, four tasks were designed to evaluate the ability of EEG-FMs to capture brain states,  AD-related cognitive function dynamics and subtle pre-symptomatic risks:

\begin{enumerate}
    \item \textbf{Resting - Eyes Open/Closed}. From the resting-state recording, consecutive non-overlapping trials of 4 seconds were extracted. 
    \item \textbf{Multi-source interference task (MSIT)}. From the MSIT task, trials of 1 second (lies within the minimum inter-stimulus interval, so trials do not overlap subsequent stimuli) anchored at the stimulus marker were extracted, labeled as congruent/low-demand vs. incongruent/high-demand. Load-related changes in executive processing are relevant to early AD, this task shows whether a model captures cognitively meaningful, task-evoked neural activity.
    \item \textbf{Sternberg Memory Task}. From the Sternberg task, trials of 3 seconds during the memorization encoding phase anchored at the stimulus marker were extracted, labeled as low-load vs. high-load. Load-related changes in working-memory processing are relevant to early AD. This task also shows whether a model captures cognitively meaningful, task-evoked neural activity. To remove artifacts not correlated with EEG, we performed Canonical Correlation Analysis ~\citep{WimDeClercq2006} cleaning of the extracted trials.
    \item \textbf{ApoE-$\varepsilon 4$ Genetic-Risk Decoding}. Using the trials extracted in the aforementioned Resting - Eyes Open/Closed task, we decode the presence or absence of $\varepsilon 4$ carrier status established by the participant’s health genetic data.
\end{enumerate}

The ~\citep{Miltiadous2023} dataset (36 AD subjects, mean age: 66.4$\pm$7.9; 29 cognitively normal, mean age: 67.9$\pm$5.4) was used to design a classification task between AD and control subjects. 

\subsection{Preprocessing}
\label{sec:preprocessing}

\paragraph{Signal Filtering / Resampling.} All recordings were processed identically with a deliberately minimal pipeline,
so that differences between models reflect their learned representations rather than hand-crafted feature engineering. For each recording we (i) retained the EEG channels, (ii) applied a band-pass filter of
0.5-45Hz and (iii) resampled to
200Hz. 

\paragraph{Channel harmonisation.}
The $128$-channel montage includes many intermediate ``$h$'' electrode positions
(e.g.\ FCC5h, AFF5h, TPP9h) that are absent from the standard $10$--$20$/$10$--$10$
vocabularies used by the spatial embeddings of EEG foundation models. We
therefore restricted all trials to the standard electrodes recognized by these
models, discarding the intermediate positions and a small number of peripheral
electrodes not present in the reference montage. This yields a consistent
channel set across trials. For each EEG-FM, a subset of electrodes was selected that is present in each model's pretraining vocabulary.

\subsection{Evaluation Protocol}

We used two evaluation settings: \emph{(i)} Full-Fine Tuning and \emph{(ii)} Head-Only. In each setting, we finetuned either the full EEG-FM (setting \emph{(i)}) or only the classification head (setting \emph{(ii)}). For both settings, we evaluated the EEG-FMs described in Section~\ref{sec:background}\footnote{BIOT's pre-trained bipolar channels were mapped to standard locations. For LEAD, the 19 channels from its pretraining montage are used.}. Each model and configuration was trained for 20 epochs to prevent overfitting
and evaluated using 10-fold subject-independent cross-validation. For
fair comparison, using the same fine-tuning setup, we performed a similar training (from scratch) process for the EEGNet model to provide a basis for comparison.

\subsection{Results}
\label{sec:results}

\begin{table*}[!h]
\centering
\caption{\textbf{Full-Tuning.} Balanced accuracy (\%, mean\,$\pm$\,s.d.\ over 10 subject-wise folds) on the four downstream tasks. Resting state eyes-open vs. eyes-closed; Cognitive load low vs. high (MSIT, Sternberg); AD vs Control classification. Best per column in \textbf{bold}. Second-best \underline{underlined}. \\}
\label{tab:pearl-bacc}
\resizebox{0.8\textwidth}{!}{%
\begin{tabular}{l|c|cc|c}
\toprule
 & \multicolumn{1}{c}{Resting State} & \multicolumn{2}{c}{Cognitive Load} & \multicolumn{1}{c}{AD Classification} \\
Model & Eyes Open-Closed & MSIT & Sternberg & AD-vs-Control \\
\midrule
EEGNet & 81.5$\pm$7.2 & \underline{71.0$\pm$2.6} & 63.5$\pm$5.2 & 71.0$\pm$4.55 \\
\midrule
LaBraM & 87.3$\pm$5.5 & 70.8$\pm$2.0 & 60.6$\pm$3.0 & \underline{72.4$\pm$13.22}\\
CBraMod & 84.5$\pm$4.8 & 70.4$\pm$2.3 & \underline{66.8$\pm$4.0}  & 70.3$\pm$11.26 \\
BIOT & 76.9$\pm$7.2 & 54.8$\pm$2.1 & 62.7$\pm$4.9 & 65.7$\pm$11.47 \\
\midrule
LEAD & 82.8$\pm$5.3 & 66.5$\pm$2.1 & 61.3$\pm$4.0 & 67.9$\pm$15.55 \\
\midrule
REVE & \textbf{87.7$\pm$5.5} & 69.2$\pm$2.9 & 65.8$\pm$4.9 & \textbf{72.7$\pm$8.11} \\
NeuroRVQ & \underline{87.6$\pm$5.8} & \textbf{71.5$\pm$2.5} & \textbf{69.8$\pm$3.1} & 69.7$\pm$15.70 \\
\bottomrule
\end{tabular}%
}
\end{table*}


Under full fine-tuning, EEG-FMs recover coarse and cognitively meaningful neural dynamics well and largely transfer across subjects. Gross brain state is the easiest target: REVE (87.7\%), NeuroRVQ (87.6\%) and LaBraM (87.3\%) all decode eyes-open vs. eyes-closed several points above the from-scratch EEGNet baseline (81.5\%), confirming that the models capture well-established wakefulness resting-state
markers. Cognitive load detection shows a clear task-dependent split. The cognitive load of the interference/executive task (MSIT) is decoded strongly: performance clusters near 70\% and NeuroRVQ (71.5\%) only marginally exceeds the supervised baseline, suggesting pretraining offers little uplift for executive-interference decoding in this cohort. The discrimination of short-term/working-memory load (Sternberg task) is markedly harder: NeuroRVQ (69.8\%), CBraMod (66.8\%) and REVE (65.8\%) lead, while the other foundation models do not exceed EEGNet. In the AD classification task, all models achieve good performance with REVE (72.7\%) and LaBraM (72.4\%). Critically, the pre-symptomatic ApoE-$\varepsilon 4$ genetic risk does not appear to be reliably decodable from the resting-state EEG signals even with full fine-tuning.

\begin{table*}[!h]
\centering
\caption{\textbf{Head-Only.} Balanced accuracy (\%, mean\,$\pm$\,s.d.\ over 10 subject-wise folds) on the four downstream tasks. Resting state eyes-open vs. eyes-closed; Cognitive load low vs. high (MSIT, Sternberg); AD vs Control classification. Best per column in \textbf{bold}. Second-best \underline{underlined}. \\}
\label{tab:pearl-bacc-probing}
\resizebox{0.8\textwidth}{!}{%
\begin{tabular}{l|c|cc|c}
\toprule
 & \multicolumn{1}{c}{Resting State} & \multicolumn{2}{c}{Cognitive Load} & \multicolumn{1}{c}{AD Classification} \\
Model & Eyes Open-Closed & MSIT & Sternberg & AD-vs-Control \\
\midrule
LaBraM & 57.7$\pm$4.6 & 57.9$\pm$3.2 & \underline{64.9$\pm$4.9} & 53.0$\pm$6.93 \\
CBraMod & \underline{82.3$\pm$5.7} & 56.1$\pm$1.8 & 66.9$\pm$3.2 & 54.1$\pm$6.71 \\
BIOT & 70.7$\pm$7.2 & 52.8$\pm$1.8 & 60.5$\pm$4.2 & 70.3$\pm$8.29 \\
\midrule
LEAD & 65.0$\pm$3.9 & 63.8$\pm$2.3 & 51.6$\pm$2.4 & \textbf{77.2$\pm$12.04} \\
\midrule
REVE & \textbf{84.2$\pm$6.7} & \textbf{68.0$\pm$2.2} & 64.7$\pm$4.5  &  69.4$\pm$10.09\\
NeuroRVQ & 81.4$\pm$5.8 & \underline{67.5$\pm$3.0} & \textbf{68.8$\pm$4.3}  &  \underline{72.5$\pm$7.64}\\
\bottomrule
\end{tabular}%
}
\end{table*}


Head-only training freezes each backbone and trains only a classification head, directly testing whether pretraining yields usable representations. The results reveal large differences in the representational quality that the full fine-tuning may mask. For resting state, REVE (84.2\%), CBraMod (82.3\%) and NeuroRVQ (81.4\%) retain strong decoding from frozen features. Cognitive load tasks show the same pattern with NeuroRVQ preserving 68.8\% (Sternberg) and 67.5\% (MSIT). In the AD classification task, LEAD leads with 77.2\%, validating its AD-specific pretraining. ApoE-$\varepsilon 4$ risk detection is random, confirming that frozen representations carry no accessible pre-symptomatic genetic information from the resting-state EEG signals. Appendix \ref{sec:stats} provides 
a statistical comparison while Appendix \ref{sec:scratch} presents a pre-trained vs from-scratch model comparison.

To investigate the spatial attribution~\citep{urban}, we computed Input x Gradient (IxG) attributions~\citep{deeplift, ancona2018inputxgradient} for the EEGNet model on the held-out subjects of each fold. Per-channel relevance was obtained by averaging the absolute attribution over time. For each class, only the top 25\% most-confident correctly classified trials were retained and averaged across folds to obtain spatial topographies. We then took the class difference to isolate the load-discriminative pattern (each class map mean-normalised prior to subtraction so the contrast reflects spatial distribution rather than overall magnitude). For EEGNet on the Sternberg task, we observed a posterior (parieto-occipital) distribution consistent with cognitive load-dependent activity during the memorisation/encoding phase of the task while for the resting-state eyes open/closed detection, the relevance localised to frontopolar channels.

\begin{figure}[!h]
    \centering
    \begin{subfigure}[t]{0.45\linewidth}
        \centering
        \includegraphics[width=0.90\linewidth]{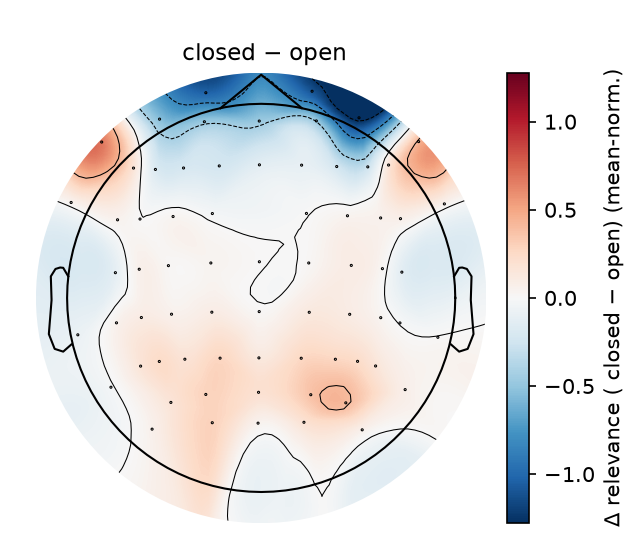}
    \end{subfigure}
    \hfill
    \begin{subfigure}[t]{0.45\linewidth}
        \centering
        \includegraphics[width=\linewidth]{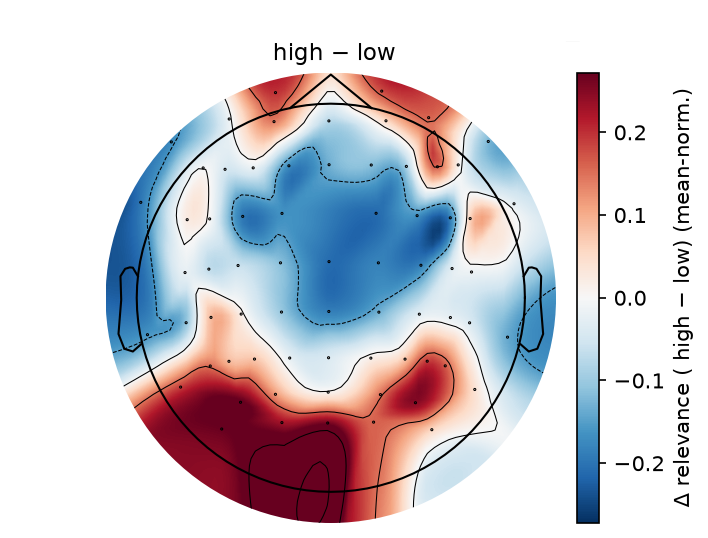}
    \end{subfigure}
    \caption{Class-averaged attribution topographic maps using IxG averaged across folds for EEGNet. (Left) Resting State Eyes Open vs. Closed. (Right) Sternberg task Low vs. High Load.}
    \label{fig:}
\end{figure}

\section{Conclusion}

In this work, we benchmarked six state-of-the-art EEG-FMs against a supervised baseline on five AD-related tasks under both full fine-tuning and head-only settings. Across settings, EEG-FMs recover cognitively meaningful, subject-transferable dynamics (robustly for resting brain state, AD classification and cognitive load in executive function and short-term/working memory, supporting their promise for scalable EEG-based biomarker development). Two findings stand out for AD. First, several general-purpose models, notably REVE and NeuroRVQ, yield linearly separable, subject-transferable descriptors without any AD-specific tuning, which suggests attractive properties for generalizability across a wider spectrum of neurodegenerative diseases. Second, LEAD leads only on AD classification under head-only, suggesting that broad large-scale pretraining already transfers well to AD-related tasks. Decoding subtle ApoE-$\varepsilon 4$ genetic risk from the resting-state information remained at chance for every model, marking it as the key open frontier. Closing this gap may rely on using a different input paradigm such as high cognitive load information (e.g. from MSIT or Sternberg task) or likely require risk-sensitive pretraining, multimodal fusion of brain and body signals and longitudinal cohorts that model decline trajectories rather than static labels.

\newpage
\bibliographystyle{plainnat}
\bibliography{neurips_2026}

\newpage
\appendix

\newpage
\section{Statistical Comparison}
\label{sec:stats}

In Table \ref{tab:statistical-test}, we show  paired significance tests (Wilcoxon signed-rank across the 10 cross-validation folds with per-fold win counts) comparing NeuroRVQ against each foundation model baseline. 

\begin{table}[!h]
\centering
\caption{Wilcoxon signed-rank tests of NeuroRVQ against each foundation model. Folds won is the number of the 10 subject-wise folds on which NeuroRVQ achieved higher balanced accuracy; $p$ is the exact two-sided $p$-value before Holm correction.}
\label{tab:statistical-test}
\footnotesize
\setlength{\tabcolsep}{3.5pt}
\begin{tabular}{l|cccc@{\hspace{20pt}}cccc}
\toprule
 & \multicolumn{2}{c}{Full fine-tuning} & \multicolumn{2}{c}{Head-Only} &
 \multicolumn{2}{c}{Full fine-tuning} & \multicolumn{2}{c}{Head-Only} \\
\cmidrule(lr){2-3}\cmidrule(lr){4-5}\cmidrule(lr){6-7}\cmidrule(lr){8-9}
NeuroRVQ vs. & Won & $p$ & Won & $p$ & Won & $p$ & Won & $p$ \\
\midrule
\multicolumn{1}{l}{} &
\multicolumn{4}{l}{\textit{Resting state: eyes open vs.\ closed}} &
\multicolumn{4}{l}{\textit{Cognitive load: Sternberg}} \\
LaBraM  & 5/10  & 0.695 & 10/10 & 0.002 & 10/10 & 0.002 & 9/10  & 0.010 \\
CBraMod & 7/10  & 0.105 & 5/10  & 0.770 & 8/10  & 0.010 & 7/10  & 0.160 \\
BIOT    & 9/10  & 0.004 & 10/10 & 0.002 & 10/10 & 0.002 & 9/10  & 0.004 \\
LEAD    & 9/10  & 0.006 & 10/10 & 0.002 & 10/10 & 0.002 & 10/10 & 0.002 \\
REVE    & 5/10  & 1.000 & 4/10  & 0.105 & 9/10  & 0.004 & 9/10  & 0.004 \\
\midrule
\multicolumn{1}{l}{} &
\multicolumn{4}{l}{\textit{Cognitive load: MSIT}} &
\multicolumn{4}{l}{\textit{AD classification: AD vs.\ Control}} \\
LaBraM  & 5/10  & 0.492 & 10/10 & 0.002 & 4/10  & 0.625 & 10/10 & 0.002 \\
CBraMod & 7/10  & 0.037 & 10/10 & 0.002 & 6/10  & 1.000 & 10/10 & 0.002 \\
BIOT    & 10/10 & 0.002 & 10/10 & 0.002 & 5/10  & 0.770 & 6/10  & 0.492 \\
LEAD    & 10/10 & 0.002 & 8/10  & 0.010 & 5/10  & 0.922 & 2/10  & 0.322 \\
REVE    & 9/10  & 0.027 & 6/10  & 0.922 & 6/10  & 0.770 & 5/10  & 0.492 \\
\bottomrule
\end{tabular}
\end{table}

In the full-finetuning setting, NeuroRVQ significantly outperformed all five foundation models on the Sternberg task, winning 8 to 10 of the 10 folds against each. On resting state and MSIT, NeuroRVQ significantly outperformed BIOT and LEAD, while its differences from LaBraM, CBraMod and REVE were small. For AD vs.\ Control, no model differed significantly from NeuroRVQ. 

In the head-only setting, NeuroRVQ significantly outperformed most foundation models on every task. It was better than LaBraM, BIOT and LEAD on resting state (10/10 folds each), better than LaBraM, CBraMod, BIOT and LEAD on MSIT, and better than LaBraM, BIOT, LEAD and REVE on Sternberg. For AD vs.\ Control, it significantly outperformed LaBraM and CBraMod, while its differences from BIOT, REVE and LEAD were not significant. LEAD, whose  pretraining includes many Alzheimer's disease related datasets, obtained the highest mean on this task. 

\section{Pre-trained vs From-Scratch Model Comparison}
\label{sec:scratch}

In Section \ref{sec:results}, Tables \ref{tab:pearl-bacc} and \ref{tab:pearl-bacc-probing} show the performance of EEG-FMs compared to the classical deep learning baseline EEGNet. In this section, we isolate the contribution of pre-training. We focus on NeuroRVQ and REVE, which are among the best-performing models in both tables (NeuroRVQ's improvements were statistically significant on the Sternberg task see Section~\ref{sec:stats}). We compare NeuroRVQ and REVE with and without pre-trained weights on the Sternberg task under both evaluation setups: full fine-tuning and head-only. As shown in Table \ref{tab:pretraining_ablation}, both pre-trained EEG-FMs achieve greater performance than their randomly initialized versions.

\begin{table}[!h]
\centering
\caption{Balanced accuracy (\%, mean over 10 subject-wise folds) of NeuroRVQ and REVE trained from randomly initialised weights (\emph{Random init.}) or from their pre-trained weights (\emph{Pre-trained}), under full fine-tuning and head-only. Values in parentheses give the gain from pre-training. EEGNet is shown as a supervised reference.}
\label{tab:pretraining_ablation}
\begin{tabular}{llcc}
\toprule
Model & Initialisation & Full fine-tuning & Head-Only \\
\midrule
EEGNet   & Random init. & 63.5 & -- \\
\midrule
NeuroRVQ & Random init. & 65.1 & 65.5 \\
         & Pre-trained  & \textbf{69.8} \,(+4.7) & \textbf{68.8} \,(+3.3) \\
\midrule
REVE     & Random init. & 57.3 & 52.8 \\
         & Pre-trained  & 65.8 \,(+8.5) & 64.8 \,(+12.0) \\
\bottomrule
\end{tabular}
\end{table}

\end{document}